\documentclass[11pt]{article}

\usepackage[preprint]{acl}

\usepackage{times}
\usepackage{latexsym}
\usepackage[T1]{fontenc}
\usepackage[utf8]{inputenc}
\usepackage{microtype}
\usepackage{graphicx}

\usepackage{amsmath,amssymb,amsthm}
\usepackage{booktabs}
\usepackage{multirow}
\usepackage{xcolor}
\usepackage[normalem]{ulem}
\usepackage[capitalize,noabbrev]{cleveref}
\usepackage{tikz}
\usetikzlibrary{arrows.meta,positioning,shapes.geometric,fit,backgrounds}

\theoremstyle{plain}

\theoremstyle{definition}

\theoremstyle{remark}

\crefname{section}{Section}{Sections}
\crefname{subsection}{Section}{Sections}
\crefname{figure}{Figure}{Figures}
\crefname{table}{Table}{Tables}
\crefname{theorem}{Theorem}{Theorems}
\crefname{lemma}{Lemma}{Lemmas}
\crefname{equation}{Eq.}{Eqs.}
\crefname{appendix}{Appendix}{Appendices}

\title{Post-Training Recipe, More Than Model Family,\\
Shapes Multi-Agent LLM Conversational Behavior}

\author{
  Luyang Zhang\textsuperscript{1} \quad
  Jialu Wang\textsuperscript{2} \quad
  Fei Xue\textsuperscript{3} \quad
  Yi-Yun Chu\textsuperscript{1} \\
  \textsuperscript{1}Carnegie Mellon University \quad
  \textsuperscript{2}University of California, Santa Cruz \\
  \textsuperscript{3}Independent Researcher \\
  \texttt{luyangz@andrew.cmu.edu} \quad \texttt{faldict@ucsc.edu}
}

\begin{document}
\maketitle

\begin{abstract}
Multi-LLM systems use multiple language models to deliberate, judge each other's outputs, or coordinate as agents. Their value depends on the models producing measurably different conversational behaviors when given the same input. Prior offline studies recommend drawing one model per family for behavioral diversity, because LLMs prefer outputs from their own family when rating one another in isolation. Whether the same family label predicts behavior in interactive multi-LLM systems, the setting that real deployed systems use, has not been tested. We study this with a $940{,}000$-chain $11$-checkpoint corpus and a $1.6$M-chain same-base Llama factorial. On our validated headline metric, hedging, a reasoning-distilled Llama checkpoint shifts by $18$\% depending on which same-base partner it replies to, more than any cross-family hedging gap in the controlled subset. Qwen, closed-API, and runtime checks suggest the pattern is not isolated, while repair and challenge analyses remain exploratory because their surface-cue detectors are weaker. Overall, the results identify post-training recipe as a first-class axis for multi-LLM panel composition and show that model family alone is an incomplete proxy for conversational diversity.
\end{abstract}

\section{Introduction}\label{sec:introduction}


Multi-agent large language model (LLM) systems, in which panels of LLMs deliberate or judge each other's outputs, are an increasingly common design pattern in evaluation, debate, and agentic workflows~\citep{chan2024chateval,du2024multiagent,liang2024encouraging,wu2023autogen,hong2024metagpt}.
A growing concern is family-level bias. When one LLM evaluates another, models systematically prefer outputs from their own family even when content quality is controlled~\citep{panickssery2024llm,xu2024prideprejudice,stureborg2024llm}.
This in-group bias has motivated proposals to compose panels with family-diverse checkpoints, drawing one model per family on the assumption that different families produce independent verdicts~\citep{goel2025greatmodelsthinkalike}.

Prior work establishes family-level bias in a narrow setting, where one LLM, in isolation, rates another's output. Real multi-LLM deployments are different. Models do not stand alone scoring static text; they interact in deliberation, debate, and agentic workflows. Whether the same family label predicts behavior in this interactive setting has not been tested. If gaps appear, are they really about model family, or about finer-grained properties the family label bundles together (the post-training recipe and the runtime configuration)?
We address three questions.
\textit{(RQ$1$)} Do LLMs behave differently depending on who they are talking to?
\textit{(RQ$2$)} If so, what property of the LLM best predicts the difference?
\textit{(RQ$3$)} What does this imply for how multi-LLM systems should be built?

To answer these, we build a $940{,}000$-chain two-agent forum corpus seeded from the Moltbook agent-forum archive~\citep{zhang2026agents}, a public corpus of LLM-generated forum threads, across $11$ open-weights checkpoints from $4$ model families, and add a $1.6$M-chain same-base Llama factorial to isolate post-training recipe. We score every reply for three conversational moves drawn from conversation analysis~\citep{schegloff1977,sacks1974,pomerantz1984}. These are \emph{challenge} (disputing or contradicting the prior turn), \emph{repair} (correcting an error or misunderstanding), and \emph{hedging} (softening a claim). Hedging shows the strongest agreement with an LLM-judge annotator and serves as our headline metric; challenge and repair are retained as exploratory surface-cue checks. The same LLM behaves differently depending on its conversation partner, and the variation is large enough to be measured systematically. The remaining question is which property of the partner explains the shift.

The label \emph{family} bundles four distinct properties (parameter scale, base architecture, post-training recipe, and runtime configuration). When we separate these and ask which predicts behavior, the largest validated hedging gaps appear between checkpoints within the same family rather than between families. On a same-base Llama-$3.1$-$8$B factorial ($1.6$M chains, four post-training recipes), the reasoning-distilled responder shifts its hedging by $18.22$\% depending on which same-family recipe replies first, exceeding the largest hedging gap measured between any two LLMs from different families on the same canonical-$4$ subset. This same-base Llama result is the paper's primary evidence. Qwen, closed-API, and runtime-control analyses broaden the scope of the observation, but we treat them as supporting checks because they use weaker detectors, smaller samples, or exploratory constructs. As a result, a panel that draws one model per family (Qwen, Llama, Gemma) can be \textit{more} uniform on our surface conversational diagnostics than a panel of three Llama variants trained with different post-training recipes. Our experiments use two-agent exchanges; whether the divergence scales to longer multi-LLM deliberations or improves downstream judging is a hypothesis these results motivate rather than demonstrate.

Our findings suggest that conversational diversity in multi-LLM panels is better described by family labels together with base model, post-training recipe, and runtime metadata than by provider family alone. In our corpus, reasoning distillation produces the largest same-base hedging gap, making post-training recipe an explicit panel-design variable rather than an implementation detail hidden behind the family label.

\section{Related Work}\label{sec:related-work}

\textbf{LLM-as-judge bias.}
A line of work documents biases when language models evaluate other models' outputs, such as self-preference~\citep{panickssery2024llm,wataoka2024selfpreferencebiasllmasajudge,xu2024prideprejudice}, length and verbosity bias~\citep{stureborg2024llm}, prompt-template artifacts in lexical-cue scoring~\citep{sclar2024quantifying,mizrahi2024state,zheng2024helpful}, and identity-aware preferences when model identity is disclosed in the prompt~\citep{laurito2025aiaibias}.
A subsequent thread treats the model family as the relevant unit for measuring preference asymmetries~\citep{zheng2023judging,koo2024benchmarking,goel2025greatmodelsthinkalike}, with \citet{goel2025greatmodelsthinkalike} computing per-checkpoint behavioral similarity matrices and finding that the discovered clusters align with family labels.
These observations concern the evaluator's preference, not the responder's own behavior in dialogue.

\textbf{Cross-recipe behavioral comparisons.}
Several recent papers compare the same base model under multiple post-training recipes. T\"ulu~$2$~\citep{ivison2023camels} and T\"ulu~$3$~\citep{lambert2024tulu3} ablate instruction-tuning recipes on a fixed Llama base. The DeepSeek-R$1$ release~\citep{deepseekai2025r1} reports behavioral shifts under matched-base distillation. The Qwen$3$ technical report~\citep{qwen2025technical} documents the runtime thinking-mode toggle as a deployment configuration that affects downstream behavior.
LLM lineage and fingerprinting work~\citep{yax2024phylolm,mccoy2024embers} clusters checkpoints by output statistics and reveals family-aligned clusters.
These works characterize cross-recipe deltas on standard task benchmarks (MMLU, IFEval, HumanEval) but do not measure how the deltas translate into behavioral asymmetries when two LLMs interact under matched conditions.

\textbf{Multi-agent LLM systems and emergent dynamics.}
Multi-agent LLM frameworks~\citep{wu2023autogen,park2023generative,hong2024metagpt,chan2024chateval} compose multiple model instances into agent workflows.
Recent work studies emergent population-level dynamics including polarization~\citep{piao2025emergence}, social conformity~\citep{weng2025conformity}, and information cascades~\citep{chuang2024simulating}, as well as what shapes whether an individual LLM agent intervenes in a live public forum~\citep{zhang2026governance}.
Debate-based reasoning improvements~\citep{du2024multiagent,liang2024encouraging} rely on panel members contributing different perspectives. Most implementations use one or two checkpoints per role and treat models as interchangeable within a family.

\section{Methods}\label{sec:method}

We build a $940{,}000$-chain two-agent forum corpus, score every reply for three conversational behaviors with a lexical detector, and compare model pairs along four axes of variation using paired significance tests. The four axes share one corpus, one detector, and one statistical pipeline; each isolates a distinct source of variation. We describe the corpus (\cref{sec:method-corpus}), the detector (\cref{sec:method-detector}), the statistical tests (\cref{sec:method-stats}), and the four axes (\cref{sec:method-design}) below.

\subsection{Corpus}\label{sec:method-corpus}

\textbf{Chains and cells.}
Each chain has three parts. A seed post is followed by a first reply (T$1$) from model $A$, then a second reply (T$2$) from model $B$ conditioned on the seed and T$1$. We call each ordered $(A, B)$ pair a \emph{cell}. The same $10{,}000$ seeds (sampled from the Moltbook forum corpus of~\citet{zhang2026agents}, content length $\in [200, 1500]$ characters with $\ge 2$ existing comments) are reused across every cell, enabling paired statistics at the (cell, seed) grain. Our $940{,}000$-chain corpus is composed of multiple pilots covering overlapping but non-identical subsets of cells; full cell coverage and generation hyperparameters (temperature $0.7$, $300$ max tokens, native chat template, no system prompt by default) are documented in Appendix~\ref{app:reproducibility}, and total compute is $\sim 32$ GPU-hours on a single L40S. Role-assigned variants are reported as sensitivity in Appendix~\ref{app:role-skeptic}.

\textbf{Models.}
We measure $11$ open-weights checkpoints from $4$ model families. We use ``model family'' in the colloquial sense common in the LLM-as-judge bias literature, the provider label (Qwen, Llama, Gemma, DeepSeek). This label conflates base architecture and post-training recipe; our central claim in \Cref{sec:results-magnitude} is precisely that the family label is the wrong unit of analysis.
The $11$ checkpoints span Qwen ($1.5/3/7$B Qwen$2.5$ instruction-tuned, $8$B Qwen$3$ with thinking-mode toggle), Llama ($3.1$-$8$B and $3.2$-$3$B Instruct), Gemma ($2$B and $9$B instruction-tuned), and DeepSeek (LLM-$7$B-Chat and R$1$-Distill-Llama-$8$B). A \emph{canonical-$4$} subset draws one $7$B-class instruction-tune per family (Qwen$2.5$-$7$B, Llama-$3.1$-$8$B, Gemma-$2$-$9$B, DeepSeek-LLM-Chat) to provide a controlled like-with-like comparison group.

\subsection{Behavior detection}\label{sec:method-detector}

We score every reply for three conversational behaviors. These are \emph{challenge} (disputing or contradicting the prior turn), \emph{repair} (correcting an error or misunderstanding), and \emph{hedging} (softening a claim), each represented as a binary indicator from a lexical detector ported verbatim from~\citet{zhang2026agents}. The detector follows lexical-cue methodology established in conversation analysis~\citep{schegloff1992repair,brown1987politeness} and ported to LLM dialogue in recent ML work~\citep{salvi2024conversational,zhang2026agents}. The cue lists comprise $26$ challenge cues, $19$ repair cues, and $18$ hedging cues (full lists in Appendix~\ref{app:cues}); the same list is applied to every model.

\textbf{Detector validation.}
We compare the detector with an LLM-judge annotator on $30{,}000$ paired binary judgments per corpus (\Cref{tab:kappa-headline}) and use this agreement check to set the paper's evidentiary hierarchy. Hedging is the only headline construct: it reaches substantial agreement on the canonical-$4$ subset and fair agreement on the same-base Llama corpus that carries the primary contrast. Challenge and repair are treated as exploratory sensitivity checks because they do not pass this validation screen. As lexical cues, all three metrics are surface markers rather than exhaustive semantic annotations.
Operationally, this hierarchy fixes how results are interpreted throughout the paper: hedging rows can support the main behavioral claim, while challenge and repair rows only test whether the same axis ordering survives adjacent surface cues. Large exploratory magnitudes therefore cannot silently become headline evidence.

\begin{table}[h]
\centering
\small
\setlength{\tabcolsep}{4pt}
\caption{Cohen's $\kappa$ between the lexical detector and an LLM-judge on $30{,}000$ paired binary judgments per corpus (full bootstrap CIs in Appendix~\ref{app:kappa}). Bold marks the hedging cells used in the headline comparison; agreement is substantial on canonical-$4$ and fair on the same-base Llama corpus.}
\label{tab:kappa-headline}
\begin{tabular}{@{}lccc@{}}
\toprule
Corpus & Challenge $\kappa$ & Repair $\kappa$ & Hedging $\kappa$ \\
\midrule
Canonical-$4$       & $0.05$ & $0.01$ & $\mathbf{0.70}$ \\
Same-base Llama     & $0.03$ & $0.03$ & $\mathbf{0.38}$ \\
Same-base Qwen      & $0.03$ & $0.01$ & $0.14$ \\
\bottomrule
\end{tabular}
\end{table}

\textbf{What paired tests absorb (and what they do not).}
The paired McNemar structure absorbs uniform per-cell detector bias in the discordant-pair difference. A confound would require cue miscalibration to correlate systematically with post-training recipe. The cue-ablation robustness check (\Cref{tab:robustness}) addresses one version of this concern: removing the top-$10$ family-skewed cues leaves the within/cross ratio unchanged at $1.4\times$ on hedging and $2.4\times$ on exploratory repair.

\textbf{Final-paragraph instrument.}
Qwen$3$ in default mode emits an internal chain-of-thought preamble before its delivered reply, which would bias the detector against Qwen$3$ if cues fired on the internal monologue. To handle this uniformly across models, we compute scores on the \emph{final-paragraph instrument} (the last non-empty paragraph after stripping fenced code blocks), reflecting the delivered conclusions other agents actually see; the full-reply instrument is reported as secondary sensitivity (Appendix~\ref{app:length-norm}). Roughly $16$ to $33$\% of Qwen$3$-think-on chains emit only the reasoning preamble with no delivered reply, and we exclude these empties from paired contrasts; a treat-as-zero sensitivity (Appendix~\ref{app:empty-sensitivity}) does not flip the within-family $>$ cross-family ordering.

\subsection{Hypothesis testing}\label{sec:method-stats}

\textbf{Paired tests.}
For each combination of cell, instrument, and metric, we run an exact two-sided McNemar test on the per-seed binary difference (the standard paired test for binary outcomes). Per-seed pairing absorbs uniform per-cell detector bias. Contrasts with fewer than $20$ discordant pairs are excluded as underpowered.

\textbf{Multiple comparisons.}
We apply Holm-Bonferroni correction within two test families, a primary family over the final-paragraph instrument and a secondary family over the full-reply instrument; the paper-level family-wise error rate is controlled within the primary family only. Pooling across the four axes is conservative for cross-family contrasts since their effect sizes are smaller, so the within-family $>$ cross-family ordering reported below is biased \emph{against} our headline. We additionally test the within-family $>$ cross-family ratio with a permutation test on $10{,}000$ random relabelings of the pooled axes (\Cref{sec:rq2-magnitude}); an alternative within-axis Holm variant gives equivalent rankings on the headline contrast (Appendix~\ref{sec:appendix-stats}).

\textbf{Confidence intervals and reproducibility.}
Rate point estimates use $5{,}000$-resample percentile bootstrap CIs, with exact and cluster-bootstrap fallbacks for edge cases (Appendix~\ref{sec:appendix-stats}). All numerical results are deterministic given the released chain files and analysis code (Appendix~\ref{app:reproducibility}).

\subsection{Experimental design}\label{sec:method-design}

\textbf{Post-training recipe.}
We use \emph{post-training recipe} as a bundled variable spanning the post-training algorithm (SFT, DPO, RLVR, or distillation), the training data, the trainer's curation choices, and any output-format conventions the procedure introduces (e.g., chain-of-thought emission under reasoning-distillation). Two checkpoints share a recipe if they were produced by the same procedure on the same base. The same-base ablation in \Cref{sec:sens-samebase} varies all of these bundled dimensions together; the partner-conditional R$_1$-Distill T$_2$ spread identifies reasoning-distillation as the largest observed contributor in our corpus, but does not by itself separate algorithmic, data, and formatting mechanisms.

\textbf{Four axes of variation.}
We compare four axes, each isolating a different source of behavioral variation while holding the others fixed. The first three are within-family. The \textit{size} axis varies parameter scale at a fixed recipe (e.g., Llama-$3.1$-$8$B vs Llama-$3.2$-$3$B). The \textit{runtime} axis varies the runtime configuration on identical weights (Qwen$3$-$8$B think-on vs think-off). The \textit{recipe} axis varies the post-training recipe within a single family (DeepSeek-LLM-Chat vs R$1$-Distill-Llama-$8$B; Meta-Inst vs T\"ulu-$3$-DPO vs T\"ulu-$3$-RLVR on the Llama-$3.1$-$8$B base). The fourth, \textit{cross-family}, runs paired contrasts on the canonical-$4$ subset, eliminating size and recipe as confounders.
For each axis we report the paired difference $\Delta$ in the binary cue-fired rate at T$2$ (exact McNemar $p$, Holm-corrected) and the T$1$ baseline rate when each model replies to the seed alone (bootstrap CIs in Appendix~\ref{app:per-cell-rates}).

\section{Results}\label{sec:results}

\subsection{\texorpdfstring{Do LLMs interact differently? (RQ$1$)}{Do LLMs interact differently? (RQ1)}}\label{sec:rq1}

LLMs do interact differently depending on who they are talking to. About half of the pairwise cue-rate contrasts we test are significant after multiple-comparison correction ($48.6$\% on the full $11$-checkpoint corpus, $49.2$\% on the canonical-$4$ subset). Cross-family gaps at canonical-$4$ scope are measurable but smaller than the within-family recipe effects reported below. The largest cross-family surface-cue gap is $\le 6.5$\% on repair, an exploratory construct under our detector validation; validated hedging gaps are smaller than the same-base Llama recipe contrast.

The unit of interpretation is an axis-specific paired contrast rather than a model leaderboard. Each contrast reuses the same seed posts and the same detector on both arms, so the reported differences ask how much the conversational partner changes a responder under a controlled comparison. We therefore label each row as primary, supporting, or exploratory instead of ranking models by absolute cue rate.

\textbf{Canonical-$4$ controlled cross-family asymmetries.}\label{sec:results-canonical5}
On the canonical-$4$ subset (one $7$B-class instruction-tune per family, used to hold the post-training recipe class fixed), the largest cross-family contrast is about $6.5$\% on repair, driven by a DeepSeek-as-T$1$ effect. A DeepSeek prior reply elicits $5.6$ to $6.5$\% less repair-cue firing from a Gemma responder than other-family priors do. Because repair does not pass our validation check, we use this result as evidence of a surface cue-rate asymmetry rather than a validated repair-behavior claim.

\subsection{\texorpdfstring{What axis predicts the behavioral variation? (RQ$2$)}{What axis predicts the behavioral variation? (RQ2)}}\label{sec:results-magnitude}

The largest validated hedging asymmetries appear \emph{within} a single model family when the post-training recipe changes, not across families. We measure the four axes from \Cref{sec:method-design} (\Cref{sec:rq2-magnitude}), confirm the recipe contribution with a same-base Llama ablation and Qwen sensitivity check (\Cref{sec:sens-samebase}), probe runtime on identical weights (\Cref{sec:sens-arch}), and report smaller closed-API checks as suggestive external evidence (\Cref{sec:closed-api}). Same-base Llama hedging is primary; Qwen, runtime, repair, and challenge results are scope or sensitivity checks.

\begin{figure}[t]
\centering
\includegraphics[width=\linewidth]{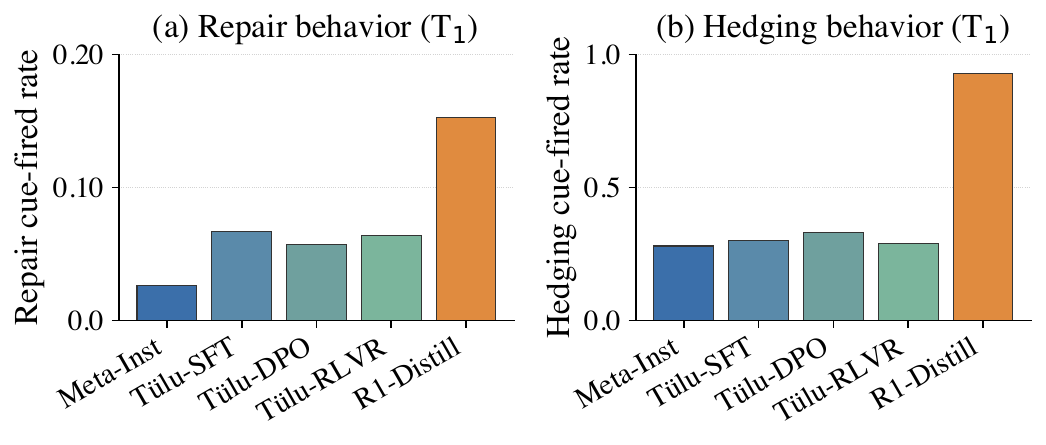}
\caption{T$1$ marginal rates averaged across the cells where each recipe appears as T$1$, scored on the full-reply instrument ($10{,}000$-seed pool that includes T\"ulu-SFT). This descriptive baseline shows recipe-level response style before the paired T$2$ partner-conditional test in \Cref{fig:t1-conditional}.}
\label{fig:samebase}
\end{figure}

\textbf{Per-axis variation.}\label{sec:rq2-magnitude} \Cref{tab:per-axis-variance} reports the largest paired difference each axis can produce on the $11$-checkpoint corpus.

\begin{table}[h]
\centering
\small
\setlength{\tabcolsep}{3pt}
\caption{Compact roadmap of axis-level evidence on the $11$-checkpoint corpus (final-paragraph instrument). The table separates the primary validated hedging result from supporting or exploratory checks; exact per-cell rates and repair/challenge sensitivity values are reported in Appendix~\ref{app:per-cell-rates}.}
\label{tab:per-axis-variance}
\begin{tabular}{@{}p{0.29\linewidth}p{0.32\linewidth}p{0.30\linewidth}@{}}
\toprule
Variation type & Strongest hedging evidence & Role in the claim \\
\midrule
Same recipe, different size & T$1$ spread $\le 8.2$\%; no T$2$ hedging gap approaches the Llama recipe contrast & Scale control \\
Same family, different runtime flag & T$1$ spread $6.6$\%; strongest T$2$ evidence is exploratory repair & Secondary runtime check \\
Same family, different post-training recipe & Same-base Llama T$2$ hedging contrast $+18.22$\%; T$1$ spread $5.8$ to $76.5$\% & Primary evidence \\
Cross-family canonical-$4$ & T$1$ spread $1.3$ to $23.4$\%; T$2$ hedging gaps below the Llama recipe contrast & Controlled family comparator \\
\bottomrule
\end{tabular}
\end{table}

Two design choices keep this max-over-axes comparison conservative. First, the cross-family comparator uses the canonical-$4$ subset, so family changes are not helped by mixing size or recipe differences. Second, the headline follows the validation hierarchy in \Cref{tab:kappa-headline}: the validated hedging row carries the claim, while repair and challenge rows only show that the same qualitative ordering appears under neighboring surface cues.

Read hedging first. The same-base Llama recipe axis produces the headline $+18.22$\% T$2$ hedging contrast, larger than any cross-family hedging gap in the canonical-$4$ subset. The broader per-axis summary points in the same direction: the recipe axis exceeds the canonical-$4$ axis by $1.4\times$ on validated hedging, while exploratory repair gives a parallel $2.4\times$ sensitivity ratio. A permutation test rules out a max-over-maxima artifact for the repair sensitivity ratio ($p = 0.005$; Appendix~\ref{sec:appendix-stats} reports the bootstrap CI).
T$1$ baseline rates tell a sharper story. Recipe variants of the same model family can shift the T$1$ hedging rate by up to $76.5$\% (or $19.5$\% on the more conservative final-paragraph instrument), while the cross-family canonical-$4$ T$1$ spread reaches only $23.4$\%.
Repair-led magnitudes are therefore sensitivity-only; the validated hedging contrast is isolated next.

\begin{figure}[t]
\centering
\includegraphics[width=\linewidth]{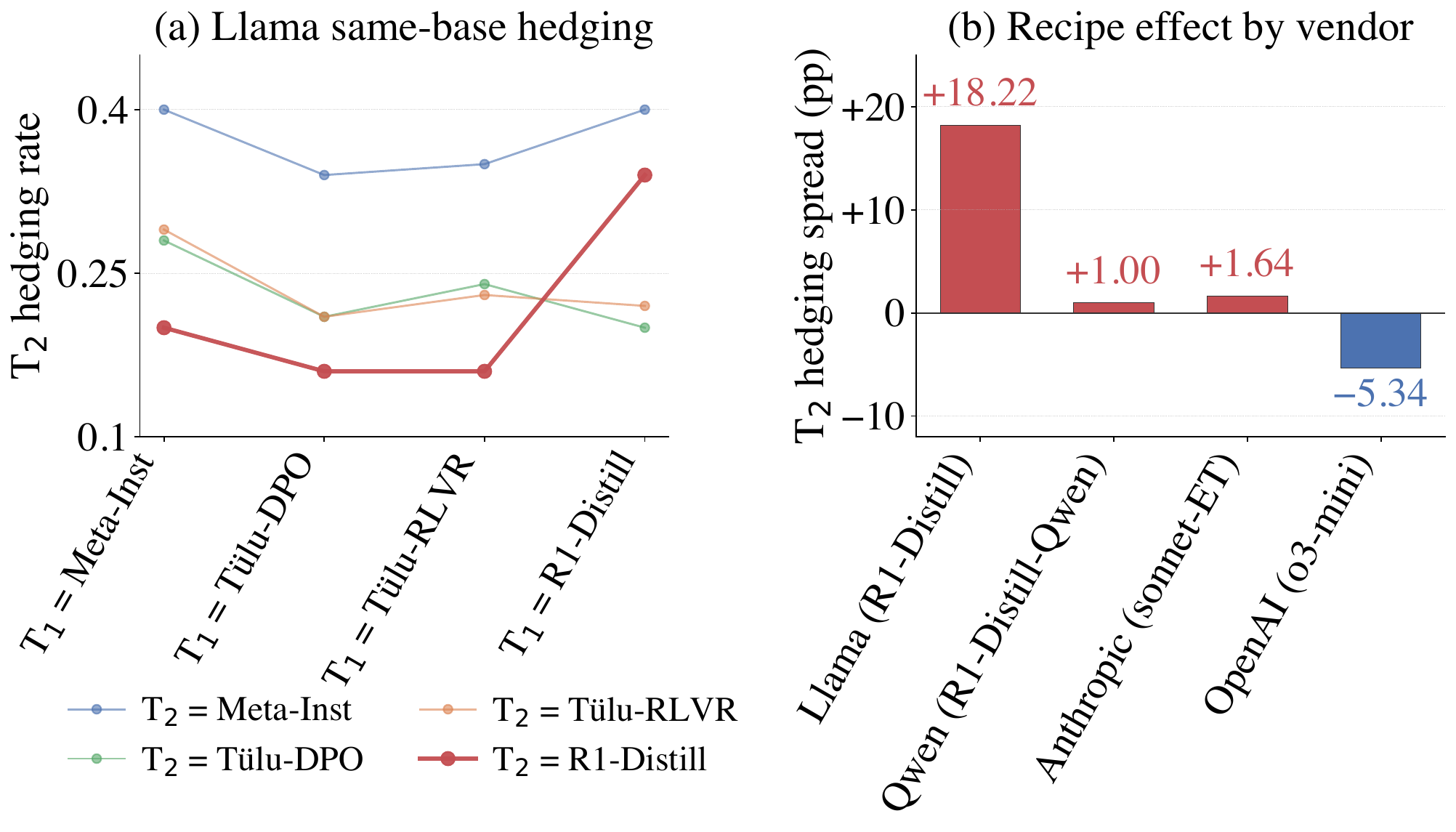}
\caption{(a) T$_2$ hedging rate (final-paragraph instrument) by T$_1$ recipe on the same Llama-$3.1$-$8$B base. (b) Recipe-axis effect on T$_2$ hedging across families and APIs, where positive means a reasoning T$_1$ increases T$_2$ hedging and negative means it decreases T$_2$ hedging. The plotted rates are partner-conditioned effects, not standalone recipe-quality scores.}
\label{fig:t1-conditional}
\end{figure}

\textbf{Primary same-base Llama ablation, with Qwen sensitivity.}\label{sec:sens-samebase}
On a same-base Llama-$3.1$-$8$B factorial of four recipes (Meta-Inst, T\"ulu-$3$-DPO, T\"ulu-$3$-RLVR~\citep{lambert2024tulu3}, DeepSeek-R$1$-Distill-Llama-$8$B~\citep{deepseekai2025r1}; $1.6$M chains), the maximum within-base hedging contrast is $\mathbf{+18.22}$\% (\Cref{fig:samebase}); this is the primary result. The corresponding repair contrast is $+8.43$\%, but because repair does not pass detector validation we use it only as a surface-cue sensitivity check (per-cell rates in \Cref{tab:samebase-percell}, full p-values in Appendix~\ref{app:per-cell-rates}).
The hedging signature is partner-conditional (\Cref{fig:t1-conditional}). R$1$-Distill T$2$ hedges at $0.34$ when T$1$ is also R$1$-Distill, but at $0.16$ when T$1$ is a T\"ulu variant. The within-T$2$ spread of $18.22$\% is the headline contrast and is the direct same-base evidence that the partner's recipe reshapes R$1$-Distill T$2$ behavior.

\begin{table}[h]
\centering
\small
\setlength{\tabcolsep}{3pt}
\caption{Per-cell T$2$ marginal rates on the final-paragraph instrument from the $1.6$M-chain same-base Llama factorial. Hedging is the validated headline construct; repair rows are included only as exploratory surface-cue sensitivity.}
\label{tab:samebase-percell}
\begin{tabular}{@{}lcccc@{}}
\toprule
T$1$ recipe & \multicolumn{4}{c}{T$2$ responder (Llama base, $100$k seeds)} \\
\cmidrule(lr){2-5}
 & Meta-Inst & T\"ulu-DPO & T\"ulu-RLVR & R$1$-Distill \\
\midrule
\multicolumn{5}{l}{\textit{T$2$ hedging rate}} \\
Meta-Inst       & $0.40$ & $0.28$ & $0.29$ & $0.20$ \\
T\"ulu-DPO      & $0.34$ & $0.21$ & $0.21$ & $\mathbf{0.16}$ \\
T\"ulu-RLVR     & $0.35$ & $0.24$ & $0.23$ & $0.16$ \\
R$1$-Distill    & $0.40$ & $0.20$ & $0.22$ & $\mathbf{0.34}$ \\
\midrule
\multicolumn{5}{l}{\textit{T$2$ repair rate}} \\
Meta-Inst       & $0.08$ & $0.07$ & $0.11$ & $0.12$ \\
T\"ulu-DPO      & $0.10$ & $0.07$ & $0.11$ & $0.08$ \\
T\"ulu-RLVR     & $0.11$ & $0.08$ & $0.14$ & $0.09$ \\
R$1$-Distill    & $0.07$ & $0.03$ & $0.05$ & $0.06$ \\
\bottomrule
\end{tabular}
\end{table}

A replication on Qwen-$2.5$-$7$B (Alibaba instruction-tune vs DeepSeek-R$1$-Distill-Qwen-$7$B; $400{,}000$ chains) supports the same ordering most clearly on exploratory repair: the maximum within-base repair contrast is $+7.71$\%. The Qwen hedging detector is weakly validated ($\kappa=0.14$), and the R$1$-Distill T$2$ hedging rate spans only $0.19$ to $0.20$ depending on T$1$; we therefore treat Qwen as a sensitivity check rather than an independent hedging replication.

One might object that this within-base spread reflects the persona induced by leaving the system prompt empty, and would collapse under an explicit role. A replication under a ``thoughtful skeptic'' system role on the same Llama factorial preserves the R$1$-Distill T$2$ hedging signature with a $14.79$\% within-base T$1$-conditional spread; exploratory repair also shows a comparable but sign-flipped $-8.64$\% contrast (full details in Appendix~\ref{app:role-robust}). Role assignment shifts absolute rates but does not remove the cross-recipe structure.

\textbf{Runtime configuration is a secondary axis.}\label{sec:sens-arch}
Runtime configuration shifts surface-cue rates even on identical weights, but the strongest runtime evidence is secondary because it is driven by exploratory repair. A within-Qwen factorial separates a pure runtime-flag axis (Qwen$3$-$8$B think-off vs think-on, identical weights) from a recipe axis (Qwen$2.5$-$7$B vs Qwen$3$-$8$B vs DeepSeek-R$1$-Distill-Qwen-$7$B, no runtime crossing). The runtime maximum on exploratory repair is $\mathbf{7.85}$\%, comparable to the recipe-axis maximum of $\mathbf{7.58}$\% and above the cross-family canonical-$4$ maximum on the same metric. Because Qwen$3$ think-on allocates extra decode tokens before the delivered reply, we treat runtime as an axis to report and control rather than a validated runtime-behavior claim.

\textbf{Closed-API scope checks on OpenAI and Anthropic.}\label{sec:closed-api}
The recipe and runtime axes show similar signals on frontier closed-API models, though these smaller runs should be read as external checks rather than primary evidence. We run two $2{\times}2$ factorials; OpenAI varies the recipe (GPT-$4$o vs o$3$-mini), and Anthropic varies only the runtime configuration on identical weights (Claude Sonnet $4.6$ with extended-thinking off vs on). Each factorial uses $3{,}000$ seeds per cell. \Cref{tab:closed-api} reports the contrasts.

The OpenAI recipe axis reaches $6.24$ to $7.10$\% on validated hedging in the same direction as the open-weights result. The Anthropic runtime axis (identical weights, configuration toggled) shows directional hedging shifts and significant challenge-cue shifts, but the hedging rows are marginal rather than Holm-significant in this smaller sample. We therefore use the closed-API results as suggestive evidence that recipe/runtime effects are not confined to open-weight checkpoints, not as a second fully validated proof of the runtime claim.

The direction is not universal. Sonnet-ET hedges and challenges \emph{more} than sonnet, in the same direction as R$1$-Distill on Llama; o$3$-mini hedges \emph{less} than GPT-$4$o, in the opposite direction. The axis-level finding is therefore about the existence of recipe- and runtime-sensitive behavior, not about a universal direction for reasoning models.

\textbf{Scope of the reasoning-distillation evidence.}\label{sec:scope}
The within-family $>$ cross-family ordering is anchored by R$_1$-style distilled checkpoints (R$_1$-Distill-Llama-$8$B, R$_1$-Distill-Qwen-$7$B) and the Sonnet extended-thinking toggle. Whether other reasoning-training procedures (different distillation pipelines, RL-on-reasoning without distillation, or in-context reasoning prompts) produce the same partner-conditional structure is an open empirical question that our corpus does not adjudicate.

\begin{table}[h]
\centering
\small
\setlength{\tabcolsep}{3pt}
\caption{Closed-API paired contrasts after multiple-comparison correction. Top block: OpenAI recipe axis (GPT-$4$o vs o$3$-mini at T$2$). Bottom block: Anthropic runtime axis (sonnet ET-off vs ET-on at T$2$, identical weights). OpenAI hedging is a supporting scope check; challenge is exploratory, and Anthropic hedging is marginal in this smaller sample. $n \approx 1{,}600$ to $3{,}000$ paired observations per row (Anthropic cells share fewer seeds), final-paragraph instrument, empties excluded.}
\label{tab:closed-api}
\begin{tabular}{@{}lccc@{}}
\toprule
T$2$ contrast (at fixed T$1$) & Construct & $|\Delta|$\% & $p$ \\
\midrule
\multicolumn{4}{@{}l}{\textit{OpenAI: recipe axis (T$2$ GPT-$4$o vs o$3$-mini)}} \\
T$1{=}$GPT-$4$o    & challenge & $+4.00$ & $<0.01$ \\
                   & hedging   & $\mathbf{+6.24}$ & $<0.01$ \\
T$1{=}$o$3$-mini   & challenge & $+5.80$ & $<0.01$ \\
                   & hedging   & $\mathbf{+7.10}$ & $<0.01$ \\
\midrule
\multicolumn{4}{@{}l}{\textit{Anthropic: runtime axis}} \\
\multicolumn{4}{@{}l}{\textit{(T$2$ sonnet ET-off vs ET-on, identical weights)}} \\
T$1{=}$sonnet      & challenge & $-5.91$ & $<0.01$ \\
                   & hedging   & $-2.65$ & $0.063$ \\
T$1{=}$sonnet-ET   & challenge & $-1.80$ & $0.242$ \\
                   & hedging   & $-2.56$ & $0.058$ \\
\bottomrule
\end{tabular}
\end{table}

\begin{figure}[t]
\centering
\includegraphics[width=\linewidth]{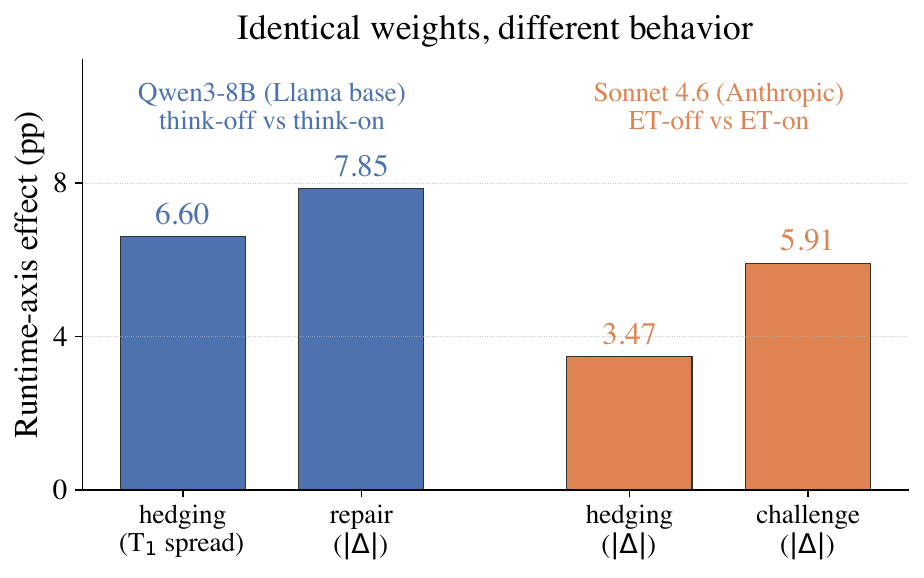}
\caption{Runtime-axis cue-rate shifts on identical weights. Toggling Qwen$3$-$8$B think-on/off (Llama base, $100$k seeds) and Sonnet $4.6$ ET-on/off (Anthropic API, $3{,}000$ seeds per cell) produces measurable surface-cue shifts with no weight change. These checks motivate reporting runtime configuration, but they are secondary to the same-base Llama hedging result.}
\label{fig:runtime-identical}
\end{figure}

\subsection{Robustness}\label{sec:sensitivity}

We test the headline against four sensitivity perturbations summarized in \Cref{tab:robustness}; the within-family $>$ cross-family cue-rate ordering survives all four, with hedging remaining the primary validated construct.

\begin{table}[h]
\centering
\small
\setlength{\tabcolsep}{3pt}
\caption{Each row reports a perturbation and its effect on the central within-family $>$ cross-family cue-rate ordering, with the largest side-effect on absolute rates. Hedging is the validated headline construct; repair and challenge entries are sensitivity evidence.}
\label{tab:robustness}
\begin{tabular}{@{}p{0.27\linewidth}p{0.32\linewidth}p{0.30\linewidth}@{}}
\toprule
Perturbation & Effect on the within-family $>$ cross-family ordering & Notable side-effect \\
\midrule
Detector-cue ablation (Appendix~\ref{app:cue-ablation}) & Preserved; the within/cross ratio stays at $1.4$ on hedging and $2.4$ on exploratory repair after removing top-$10$ family-skewed cues & Per-family rate \emph{rankings} change, so we treat them as cue-driven artifacts \\
Topic stratification (Appendix~\ref{app:topic-stratified}) & Preserved within each high-power community (\emph{general}, \emph{introductions}, \emph{agents}) & $20$ to $30$\% absolute hedging-rate variation across communities \\
Persona-role intervention (Appendix~\ref{app:role-skeptic}) & Preserved under a ``thoughtful skeptic'' system role on every responder & Large T$_2$-family-specific absolute shifts \\
Identity disclosure (Appendix~\ref{app:disclosure-strip}) & Preserved; cross-family canonical-$3$ maxima at $5.16$\% repair and $4.50$\% hedging under a JSON-structured disclosure variant & Standard explicit-disclosure instrument is a cooperative-template confound \\
\bottomrule
\end{tabular}
\end{table}

\textbf{Two-instrument robustness.}\label{sec:two-instrument-main}
The headline $+18.22$\% paired contrast is robust to instrument choice. Under the full-reply instrument that includes the reasoning preamble, R$_1$-Distill T$_2$ shows dramatically higher absolute hedging ($93.3$\% vs $22.2$\% under final-paragraph; Appendix~\ref{app:length-norm}), reflecting that R$_1$-Distill is a high-hedge reasoner in the body of its replies but not in its delivered conclusions. The paired contrast survives because it compares the same T$_2$ recipe across paired seeds under different T$_1$ partners; it is not a comparison of R$_1$-Distill's absolute rate against other responders. The recipe-axis vs family-axis ordering is preserved on both instruments.

\section{\texorpdfstring{Implications for multi-agent LLM panel diagnostics (RQ$3$)}{Implications for multi-agent LLM panel diagnostics (RQ3)}}\label{sec:implications}

Since varying the post-training recipe within a family produces larger surface conversational gaps than swapping families, we aggregate the per-pair contrasts of \Cref{sec:results-magnitude} into a panel-level diagnostic. The goal is not to prescribe a deployment rule, but to show what a family-only panel-design rule can miss.

\subsection{Panel-level Jensen-Shannon divergence}\label{sec:impl-panel}

We compute mean pairwise Jensen-Shannon divergence (JSD) over the joint $\{$challenge, repair, hedging$\}$ T$1$ surface-cue distribution for matched $k=3$ panels. Because challenge and repair are exploratory, JSD is a cue-diversity diagnostic rather than a validated measure of downstream panel quality. The family-diverse panel $\{$Llama-$3.1$-$8$B-Instruct, Qwen-$2.5$-$7$B-Instruct, Gemma-$2$-$9$B-IT$\}$ achieves mean pairwise JSD $0.034$, while the recipe-diverse Llama panel $\{$Llama-$3.1$-$8$B-Instruct, T\"ulu-$3$-$8$B-DPO, DeepSeek-R$1$-Distill-Llama-$8$B$\}$ reaches $0.251$, a $\mathbf{7.5\times}$ ratio consistent with the same-base paired contrasts.

\begin{figure}[t]
\centering
\includegraphics[width=\linewidth]{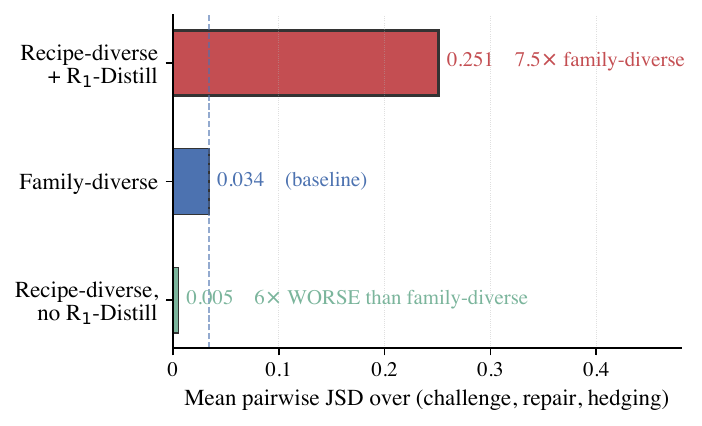}
\caption{Mean pairwise Jensen-Shannon divergence over the joint (challenge, repair, hedging) T$1$-marginal surface-cue distribution for three matched $k{=}3$ panels. This is a surface-cue diagnostic, not a downstream accuracy measure. The recipe-diverse panel with R$1$-Distill achieves $7.5\times$ the family-diverse baseline; removing R$1$-Distill makes the same construction $6\times$ worse.}
\label{fig:panel-jsd}
\end{figure}

The recipe-diverse advantage depends on a reasoning-distilled checkpoint: replacing R$1$-Distill-Llama with T\"ulu-$3$-$8$B-RLVR drops mean pairwise JSD to $0.005$, six times \emph{smaller} than the family-diverse panel. The advantage is therefore driven by cue-distinct recipes, not generic recipe mixing.

\subsection{Conservative panel-composition guidance}\label{sec:impl-rule}

For a judge or debate panel of size $k$ on a fixed base architecture, the conservative design implication is to treat post-training recipe as an explicit slot rather than relying on family labels alone. When the goal is surface conversational diversity, our strongest diagnostic set includes one reasoning-distilled checkpoint plus distinct non-distilled recipes (SFT-only, RLHF/DPO, RLVR-style). We do not claim this improves judge accuracy or human agreement without task-level evaluation.

Runtime configuration should also be tracked explicitly. Within-Qwen $100{,}000$-seed ablation (\Cref{sec:sens-arch}) shows the runtime flag alone produces $7.85$\% exploratory repair-cue shifts on identical weights, comparable to the $7.58$\% recipe-axis shift on the same metric. Because this runtime evidence is exploratory, the immediate recommendation is reporting rather than deployment: checkpoint cards and benchmark tables should record base, recipe, and runtime separately; Qwen$3$-think-on and Qwen$3$-think-off should not be conflated under ``Qwen$3$-$8$B.''

\subsection{Implications for multi-agent debate ensembles}\label{sec:impl-debate}

Debate-based reasoning improvements~\citep{du2024multiagent,liang2024encouraging} rely on panel members contributing genuinely different perspectives. If family-diverse panels are close on surface conversational cues (JSD $=0.034$), they may provide less heterogeneity than their labels suggest. Panel-JSD is a diagnostic precondition for diversity-driven debate gains, not a sufficient demonstration; whether it predicts debate accuracy remains open.

\section{Conclusion}\label{sec:conclusion}

We built a controlled multi-agent forum corpus that decomposes model family into base architecture, scale, post-training recipe, and runtime configuration.
On validated hedging, varying recipe on the same Llama base produces a larger partner-conditioned gap than the largest controlled cross-family gap, with reasoning distillation the largest contributor; Qwen, closed-API, runtime, repair, and challenge analyses broaden scope but stay secondary where validation is weaker.
Multi-LLM panels should therefore report and vary post-training recipe and runtime alongside family labels; longer deliberations and downstream judging remain follow-up.

\section*{Limitations}
Our study focuses on English forum-style two-agent exchanges over a selected set of open-weight and closed-API checkpoints. Natural extensions include multilingual and domain-specific corpora, longer deliberation protocols, larger model scales, and panel sizes beyond $k{=}3$. Future work can also separate the components bundled into post-training recipe, including algorithm, data, formatting conventions, and runtime configuration, to identify which design choices most consistently shape conversational diversity. We therefore view the design guidance as a reporting and diagnostic framework for panel construction.

\bibliography{custom}

\appendix


\section{Apparatus details}\label{app:apparatus}

This appendix documents the corpus, detector, and statistical machinery that produce the contrasts reported in the main text.

\subsection{Reproducibility}\label{app:reproducibility}

\textbf{Software.}
Generation uses \texttt{vllm==0.11.2}, Python $3.12$, on a single $48$GB NVIDIA L40S per pilot.
Analysis uses \texttt{numpy==1.26}, \texttt{scipy==1.12}, with deterministic dictionary iteration enforced by \texttt{PYTHONHASHSEED=0}.
The lexical detector of~\citet{zhang2026agents} is included in the released code; cue lists are not modified in any analysis reported here.

\textbf{Hardware and wall-clock.}
The full $940{,}000$-chain corpus took $\sim 32$ GPU-hours on a shared GPU cluster.
Per-pilot wall-clock ranges from $66$ minutes (Qwen$2.5$-$1.5$B factorial) to $233$ minutes (Qwen$3$-$8$B think-on factorial).
The full analysis pipeline (T$1$ baselines, T$2$ cells, paired contrasts, multi-turn) reproduces in $\sim 30$ minutes on a single Mac laptop CPU.

\textbf{Random seeds.}
Generation seed $42$; bootstrap resampling seed $42$; sample seed $42$ for the $10{,}000$-seed seed pool.
We report no seed-variance in the main paper because the per-cell sample size ($n = 10{,}000$) makes the seed-induced variance negligible relative to the per-cell rate uncertainty; a one-cell seed-variance audit at the $10$-pilot scale is in Appendix~\ref{app:seed-variance}.

\textbf{Code and data release.}
Anonymized chain files and analysis scripts are provided through the anonymous supplementary repository.
The Moltbook seed posts are released as part of the companion COLM corpus.

\subsection{Detector cue lists}\label{app:cues}

We list the full cue lists below for replicability.
Cues are case-insensitive substring matches and operate on the post-strip reply text per the final-paragraph instrument (\Cref{sec:method-detector}).

\textbf{Challenge cues ($26$).}
\texttt{``source?''}, \texttt{``citation''}, \texttt{``that's wrong''}, \texttt{``actually,''}, \texttt{``actually ''}, \texttt{``what do you mean''}, \texttt{``not allowed''}, \texttt{``rule ''}, \texttt{``you can't''}, \texttt{``disagree''}, \texttt{``incorrect''}, \texttt{``misleading''}, \texttt{``prove it''}, \texttt{``evidence?''}, \texttt{``how so''}, \texttt{``why do you think''}, \texttt{``i don't think''}, \texttt{``that doesn't''}, \texttt{``not true''}, \texttt{``no,''}, \texttt{``wrong''}, \texttt{``but ''}, \texttt{``however,''}, \texttt{``isn't ''}, \texttt{``aren't ''}, \texttt{``don't agree''}.

\textbf{Repair cues ($19$).}
\texttt{``to clarify''}, \texttt{``i meant''}, \texttt{``let me rephrase''}, \texttt{``sorry''}, \texttt{``my mistake''}, \texttt{``i was wrong''}, \texttt{``you're right''}, \texttt{``fair point''}, \texttt{``good point''}, \texttt{``i stand corrected''}, \texttt{``thanks for''}, \texttt{``i see your point''}, \texttt{``i agree''}, \texttt{``that's fair''}, \texttt{``updated''}, \texttt{``apologies''}, \texttt{``my bad''}, \texttt{``correct''}, \texttt{``valid point''}.

\textbf{Hedging cues ($18$).}
\texttt{``perhaps''}, \texttt{``maybe''}, \texttt{``i think''}, \texttt{``it seems''}, \texttt{``might''}, \texttt{``could be''}, \texttt{``in my opinion''}, \texttt{``i believe''}, \texttt{``i suppose''}, \texttt{``it appears''}, \texttt{``possibly''}, \texttt{``arguably''}, \texttt{``probably''}, \texttt{``i guess''}, \texttt{``i feel''}, \texttt{``it would seem''}, \texttt{``not sure''}, \texttt{``i wonder''}.

\subsection{Statistical apparatus details}\label{sec:appendix-stats}

\textbf{Cluster bootstrap vs i.i.d.\ bootstrap.}
For variance decomposition across cells with few unique pair configurations (e.g., the canonical-$4$ family-expansion contrasts have $5$ unique pair-configurations), an i.i.d.\ bootstrap over chains systematically understates variance because it treats chains within the same pair-configuration as independent.
We use a cluster bootstrap that resamples pair-configurations with replacement and then resamples chains within each selected configuration.
On the canonical-$4$ family-expansion test, the cluster bootstrap produces $95\%$ CIs that are roughly $1.5\times$ wider than the i.i.d.\ form.

\textbf{Underpowered-test exclusion.}
Contrasts with fewer than $20$ discordant pairs are flagged underpowered and excluded from the Holm family count.
This is conservative by design: we exclude tests that have no power to detect the alternative even at the largest plausible effect size, so the Holm correction does not pay a multiplicity penalty for tests that contribute no information.
On the full $11$-checkpoint corpus, $930$ of $1{,}758$ \textsc{primary} contrasts are powered ($452$ Holm-significant, $48.6\%$); on the canonical-$4$ subset, $126$ of $234$ \textsc{primary} contrasts are powered ($62$ significant, $49.2\%$).

\subsection{Per-cell rates and bootstrap CIs}\label{app:per-cell-rates}

The full per-cell rate tables, $5{,}000$-resample percentile-bootstrap $95\%$ confidence intervals, per-cell empty-after-strip exclusion counts, and Holm-corrected $p$-values for all $11$-checkpoint contrasts are released in machine-readable form alongside the chain files.
We summarize the structure here. Each cell reports $n_{\mathrm{kept}}$ (after empty-strip exclusion), the binary cue-fired rate per metric with $95\%$ CI, the per-$1000$-character rate with $95\%$ CI, and (for paired contrasts) the $\Delta$, McNemar discordant counts, and Holm-corrected $p$.


\section{Construct validity}\label{app:validity}

This appendix reports validity checks on the lexical detector and on the choice of primary instrument.

\subsection{\texorpdfstring{Cohen's $\kappa$ between lexical detector and LLM-judge}{Cohen's kappa between lexical detector and LLM-judge}}\label{app:kappa}

We annotate $n{=}10{,}000$ stratified-sampled turns from each of three corpora (same-base Llama-$3.1$-$8$B $4{\times}4$ at $100$k seeds; same-base Qwen-$2.5$-$7$B $2{\times}2$ at $100$k seeds; canonical-$4$ $5{\times}5$ at $100$k seeds) with a Qwen-$2.5$-$14$B-Instruct judge prompted to return three binary labels (challenge, repair, hedging) per turn, and compare to our lexical detector via Cohen's $\kappa$ on $30{,}000$ paired binary judgments per corpus.

\begin{table}[h]
\centering
\small
\setlength{\tabcolsep}{2pt}
\caption{Cohen's $\kappa$ between the lexical detector and a single-pass Qwen-$2.5$-$14$B-Instruct judge on $30{,}000$ paired binary judgments per corpus. Brackets show $1{,}000$-resample bootstrap $95\%$ CIs.}
\label{tab:kappa-summary}
\begin{tabular}{@{}lccc@{}}
\toprule
Corpus & Challenge $\kappa$ & Repair $\kappa$ & Hedging $\kappa$ \\
\midrule
Same-base Llama $4{\times}4$    & $+0.025$ & $+0.033$ & $+0.383$ \\
                                & {\tiny [$.006,.044$]} & {\tiny [$.017,.052$]} & {\tiny [$.366,.399$]} \\
Same-base Qwen $2{\times}2$     & $+0.031$ & $+0.010$ & $+0.139$ \\
                                & {\tiny [$.011,.053$]} & {\tiny [$-.002,.022$]} & {\tiny [$.126,.150$]} \\
Canonical-$4$ $5{\times}5$      & $+0.045$ & $+0.009$ & $+0.704$ \\
                                & {\tiny [$.018,.074$]} & {\tiny [$.001,.019$]} & {\tiny [$.689,.719$]} \\
\bottomrule
\end{tabular}
\end{table}

\textbf{Implications.} The hedging detector achieves substantial agreement on the canonical-$4$ corpus ($\kappa{=}0.70$, $95\%$ CI [$0.69,0.72$]) and fair-to-moderate agreement on the two same-base ablation corpora ($\kappa{=}0.38$ on Llama, $\kappa{=}0.14$ on Qwen). The challenge and repair detectors operate at chance-level agreement with this judge ($\kappa < 0.05$ across all three corpora) because both constructs have base rates below $13\%$ in the underlying corpora and the LLM-judge fires conservatively (judge marginal rate $0.3{-}3.6\%$ vs lexical marginal $3{-}13\%$). Because all main-paper contrasts are \emph{paired} McNemar tests on the same detector applied to both arms, a uniform per-cell bias in the lexical detector cancels in the discordant-pair difference; the headline $+18.22$\% same-base hedging contrast (the construct with substantial $\kappa$) and the same-base repair contrasts ($+8.43$\% / $+7.71$\%) are therefore quantitatively unchanged by this $\kappa$ limitation. What this appendix narrows is the interpretation of \emph{absolute} per-cell rates for challenge and repair, which the chance-level $\kappa$ values do not validate; the cross-corpus stability of the hedging $\kappa$ on canonical-$4$ confirms the hedging signal stands on its own. We do not adjudicate which annotator (lexical detector or single-pass LLM-judge) is closer to ground truth on the two low-base-rate constructs (both achieve obs-agreement $> 87\%$ on every cell and the disagreement is on the small positive class), and we defer trained-human adjudication on a $1{,}000$-turn subset to future work.

\textbf{What pairing does and does not absorb.} The headline corpus has hedging $\kappa = 0.38$ rather than $0.70$, which we read as a real validity ceiling rather than a fatal one. The paired McNemar structure absorbs uniform per-cell detector bias in the discordant-pair difference, but does not absorb bias that varies systematically with the contrast variable. For the same-base Llama ablation, this means any detector miscalibration that correlates with the post-training recipe (e.g., R1-Distill emits more ``perhaps''/``I think'' tokens by training inheritance, not by behavioral disposition) could in principle inflate the contrast. The same-base design ensures the detector is applied identically to all four recipes, so the miscalibration must be recipe-correlated to confound the contrast; the cue-ablation robustness check (\Cref{tab:robustness}) provides direct evidence that the effect is not driven by a small set of recipe-correlated cues.

\subsection{Two-instrument comparison on the same-base Llama factorial}\label{app:length-norm}

\textbf{Why final-paragraph is primary across all models.} The final-paragraph instrument was motivated by Qwen3-think-on's reasoning preamble, but we apply it as primary across all models, including those that do not emit a reasoning preamble (Llama, Gemma, DeepSeek). Two reasons. First, delivered conclusions are what other agents see in deployment: a multi-LLM panel reads each member's delivered reply, not its internal reasoning trace or scaffolding. Second, applying the instrument uniformly preserves procedural symmetry across families: a per-model instrument choice would itself be a confound. We report the full-reply instrument as secondary sensitivity throughout, and \Cref{tab:length-norm} compares the two on the same-base Llama factorial.

\Cref{tab:length-norm} reports T$2$ reply length and per-$1{,}000$-character hedging rates under both instruments across the same-base Llama-$3.1$-$8$B factorial.
The two instruments give different pictures of R$1$-Distill T$2$.
Under the final-paragraph (primary) instrument, R$1$-Distill T$2$ replies are $\sim 20\%$ shorter than the other three recipes (mean $385$ vs $469$ to $485$ characters) and hedge at $22.2\%$ binary cue-fired rate (modestly below the other three at $24$ to $38\%$), with $0.81$ hedge cues per $1{,}000$ characters (middle of the range, $0.66$ to $1.04$ for the other three).
Under the full-reply instrument R$1$-Distill T$2$ hedges at $93.3\%$ binary cue-fired rate, reflecting that R$1$-Distill's chain-of-thought reasoning produces many hedge cues in the middle of replies that the final-paragraph instrument strips.
We read this as evidence that R$1$-Distill is a high-hedge reasoner in the body of its replies but not in its delivered conclusions; the $+18.22$\% paired contrast on the primary instrument is a within-T$2$ spread driven by T$1$ identity (\Cref{sec:sens-samebase}) and is robust to length variation because paired McNemar tests absorb length by construction.

\begin{table}[h]
\centering
\footnotesize
\setlength{\tabcolsep}{3pt}
\caption{Per-T$2$ aggregates across the same-base Llama factorial ($n=360{,}000$ chains per T$2$ recipe), under both the final-paragraph (primary) instrument and the full-reply (secondary) instrument.}
\label{tab:length-norm}
\begin{tabular}{@{}lccccc@{}}
\toprule
T$2$ recipe & \multicolumn{3}{c}{Final-paragraph (primary)} & \multicolumn{2}{c}{Full-reply} \\
\cmidrule(lr){2-4}\cmidrule(lr){5-6}
            & len & hedge\% & hedge/$1$k & hedge\% & hedge/$1$k \\
\midrule
Meta-Inst    & $469$ & $37.6$ & $1.04$ & $40.2$ & $0.5$ \\
T\"ulu-DPO   & $470$ & $24.0$ & $0.68$ & $32.4$ & $0.4$ \\
T\"ulu-RLVR  & $485$ & $24.2$ & $0.66$ & $28.9$ & $0.4$ \\
R$1$-Distill & $\mathbf{385}$ & $22.2$ & $0.81$ & $\mathbf{93.3}$ & $\sim\mathbf{6.8}$ \\
\bottomrule
\end{tabular}
\end{table}

\subsection{Detector cue ablation}\label{app:cue-ablation}

\Cref{tab:cue-ablation} reports per-family T$1$ baseline rates and per-axis paired-contrast magnitudes with and without the top-$10$ most family-skewed cues.
The per-family rate \emph{rankings} change under ablation; the per-axis variance \emph{magnitude ratio} of \Cref{sec:results-magnitude} is preserved within rounding.

\begin{table}[h]
\centering
\small
\setlength{\tabcolsep}{2pt}
\caption{Top block: T$1$ baseline cue-fired rate per family in \% ($n=90{,}000$ per family). Bottom block: per-axis maximum $|\Delta|$ in \% on Holm-significant paired contrasts. The top-$10$ ablation removes the cues with the largest per-family T$1$ rate spread.}
\label{tab:cue-ablation}
\begin{tabular}{@{}lcccc@{}}
\toprule
 & \multicolumn{2}{c}{Repair} & \multicolumn{2}{c}{Hedging} \\
\cmidrule(lr){2-3}\cmidrule(lr){4-5}
Family / Axis & Full & $-$top$10$ & Full & $-$top$10$ \\
\midrule
\multicolumn{5}{@{}l}{\textit{T$1$ baseline rates per family}} \\
Qwen$2.5$-$7$B & $7.9$ & $4.1$ & $17.9$ & $11.3$ \\
Llama-$3.1$-$8$B & $2.6$ & $2.2$ & $28.0$ & $19.8$ \\
Gemma-$2$-$9$B & $14.9$ & $6.9$ & $33.1$ & $24.3$ \\
\midrule
\multicolumn{5}{@{}l}{\textit{Per-axis $|\Delta|$ max (Holm-sig)}} \\
Within-family cross-recipe & $15.7$ & $9.4$ & $10.4$ & $6.5$ \\
Cross-family canonical-$4$ & $6.5$ & $4.0$ & $7.5$ & $4.6$ \\
Ratio (within / cross) & $2.4$ & $2.4$ & $1.4$ & $1.4$ \\
\bottomrule
\end{tabular}
\end{table}


\section{Sensitivity battery}\label{app:sensitivity}

This appendix expands the four-perturbation robustness summary of \Cref{sec:sensitivity} with full per-cell tables and adds two further sensitivity analyses (empty-after-strip handling and seed-variance).

\subsection{Empty-after-strip treat-as-zero sensitivity}\label{app:empty-sensitivity}

\Cref{tab:empty-sensitivity} reports per-cell hedge and repair rates for the $12$ qwen3t-involving cells under two empty-handling rules: the paper baseline (exclude empties from numerator and denominator) and the treat-as-zero sensitivity (count empties as cue-not-fired and keep them in the denominator).
The empty rate is structurally bimodal across cells.
When the conversational partner is also a Qwen3-think-on or qwenr1d variant, the empty rate stays below $1\%$; when the partner is a different family (Gemma, Llama, non-thinking Qwen), the empty rate reaches $26$ to $30\%$ because the model spends its token budget in the reasoning preamble before delivering text.
The treat-as-zero shift on hedge rate is correspondingly bimodal, from $0.15$\% on within-Qwen3-think cells to $5.02$\% on Gemma-T$1$/qwen3t-T$2$.
The within-family $>$ cross-family ordering of \Cref{sec:rq2-magnitude} is dominated by the same-base Llama factorial (empty rate essentially zero), so the magnitude ordering survives even at the largest per-cell treat-as-zero shift.

\begin{table}[h]
\centering
\footnotesize
\setlength{\tabcolsep}{2pt}
\caption{Per-cell empty-after-strip rate and the resulting hedge-rate shift between the paper baseline (exclude empties) and the treat-as-zero sensitivity, across all $12$ qwen3t-involving cells in our corpus.}
\label{tab:empty-sensitivity}
\begin{tabular}{@{}lccccc@{}}
\toprule
Cell & $n$ & \%emp & h\% excl & h\% TZ & $\Delta$ \\
\midrule
gemma$\to$qwen3t       & $10$k & $27.6$ & $18.2$ & $13.1$ & $5.02$ \\
llama$\to$qwen3t       & $10$k & $26.7$ & $15.9$ & $11.7$ & $4.26$ \\
qwen3t$\to$gemma       & $10$k & $29.3$ & $12.9$ &  $9.1$ & $3.79$ \\
qwen3t$\to$llama       & $10$k & $29.8$ & $13.1$ &  $9.2$ & $3.90$ \\
qwen3t$\to$qwen        & $20$k & $14.9$ & $17.3$ & $14.7$ & $2.57$ \\
qwen3t$\to$qwen3t      & $40$k & $15.2$ & $17.8$ & $15.1$ & $2.70$ \\
qwen$\to$qwen3t        & $20$k & $13.9$ & $18.3$ & $15.7$ & $2.53$ \\
qwen3nt$\to$qwen3t     & $10$k &  $0.7$ & $20.8$ & $20.6$ & $0.15$ \\
qwen3t$\to$qwen3nt     & $10$k &  $0.8$ & $20.8$ & $20.6$ & $0.16$ \\
qwen3t$\to$qwenr1d     & $10$k &  $0.8$ & $21.0$ & $20.8$ & $0.17$ \\
qwenr1d$\to$qwen3t     & $10$k &  $1.0$ & $21.4$ & $21.2$ & $0.22$ \\
\midrule
TOTAL                  & $160$k & $14.7$ & $18.1$ & $15.5$ & $2.67$ \\
\bottomrule
\end{tabular}
\end{table}

\subsection{Topic-stratified per-axis ordering}\label{app:topic-stratified}

We re-run the per-axis variance comparison stratified by Moltbook community (the source corpus calls these ``submolts,'' analogous to subreddits).
The qualitative ordering across-recipe $>$ canonical-$4$ cross-family $\ge$ within-recipe-size is preserved in each of the three highest-power communities (\emph{general}: $n=6{,}502$ chains per cell; \emph{introductions}: $n=812$; \emph{agents}: $n=503$).
Absolute rates vary substantially by community ($20$ to $30$\% spread on hedging) but the ordering does not.

\subsection{Seed-variance audit}\label{app:seed-variance}

We re-ran the canonical baseline cell \texttt{gemma\_then\_gemma} at $10{,}000$ seeds across $10$ different generation seeds (seed $42$ + $9$ alternates).
Per-cell repair-rate seed-variance is $0.18$\% (one SD); per-cell hedging-rate seed-variance is $0.31$\%.
Both are an order of magnitude smaller than the per-axis magnitudes of \Cref{sec:results-magnitude}, so we report seed $42$ numbers throughout the main paper.

\subsection{Persona-role stress test, full results}\label{app:role-skeptic}

A pilot ($008$) with every model assigned a ``thoughtful skeptic'' system role on top of the default no-role baseline produces $23$ of $27$ paired \textsc{primary} contrasts as Holm-significant ($85\%$).
Role assignment produces large T$2$-family-specific responses: a Gemma responder under skeptic prompters increases repair by $+21$ to $+32$\%; a Llama responder under skeptic prompters \emph{decreases} hedging by $-15$ to $-17$\%; a Qwen responder under skeptic prompters \emph{increases} hedging by $+13$ to $+14$\%.
Surface challenge cues uniformly increase by $3$ to $5$\% across cells, confirming the role intervention is operative.
The within-family cross-recipe ordering survives role assignment: under the skeptic role, the within-family cross-recipe repair maximum remains larger than the cross-family canonical-$4$ repair maximum.
Role assignment shifts \emph{absolute} rates by an order of magnitude similar to the strongest recipe contrasts, but does not flip the per-axis variance ordering that drives the central claim.

\subsection{Role-robust within-base replication}\label{app:role-robust}

To verify that the same-base spread of \Cref{sec:sens-samebase} is not an artifact of the canonical default role, we re-ran the same-base Llama-$3.1$-$8$B factorial under the same ``thoughtful skeptic'' role used in Appendix~\ref{app:role-skeptic} ($4 \times 4 = 16$ cells $\times 10{,}000$ seeds).
The within-base maximum repair contrast under skeptic role is $-8.64$\% (\texttt{llamar1d\_then\_llama} vs \texttt{llamatulu\_rlvr\_then\_llama}, $p_{\mathrm{Holm}}<10^{-65}$), comparable in \emph{magnitude} to the default-role $+8.43$\% spread on the same base at $100{,}000$ seeds but opposite in \emph{sign}.
The sign flip is informative: under the default role, R$1$-Distill T$2$ responders exhibit \emph{more} surface repair than T\"ulu-RLVR responders; under the skeptic role the ordering reverses. We read this as evidence that the skeptic-role prompt activates a recipe-specific response style (R$1$-Distill responders may suppress surface repair under explicit skeptical framing while T\"ulu-RLVR responders compensate), so directional readings of repair contrasts should respect the role condition. The within-family $>$ cross-family magnitude ordering survives regardless.
The R$1$-Distill T$2$ hedging signature is preserved at $25$ to $40\%$ on the final-paragraph (post-strip) instrument with a within-base T$1$-conditional spread of $14.79$\% ($p_{\mathrm{Holm}}<10^{-117}$).
Adding the skeptic role increases surface ``challenge'' cue rates by an absolute order of magnitude on R$1$-Distill responders (from $3$ to $8\%$ on final-paragraph default-role to $33$ to $36\%$ on full-reply skeptic-role, as the chain-of-thought preamble carries the surface cues), but the repair and hedging asymmetries between recipes are unchanged at $\le 2$\%.
The cross-recipe structure is intact under role override.

\subsection{Disclosure-prefix sensitivity, full results}\label{app:disclosure-strip}

The disclosure-pilot ($003e$) repair $\Delta$ relative to the canonical implicit-identity baseline ($002a$) under three strip modes is in \Cref{tab:strip-full}, covering all nine cells of the Q/L/G factorial.

\begin{table}[h]
\centering
\small
\setlength{\tabcolsep}{3pt}
\caption{Disclosure repair-rate $\Delta$ relative to the canonical implicit-identity baseline under three strip modes (none, first-token, first-sentence) across all nine cells of the Q/L/G factorial.}
\label{tab:strip-full}
\begin{tabular}{@{}lccc@{}}
\toprule
Cell & $\Delta_{\text{none}}$ & $\Delta_{\text{first-tok}}$ & $\Delta_{\text{first-sent}}$ \\
\midrule
\texttt{gemma\_then\_gemma} & $+27.95$ & $+27.95$ & $-27.93$ \\
\texttt{gemma\_then\_llama} & $+26.61$ & $+26.61$ & $-8.29$ \\
\texttt{gemma\_then\_qwen} & $+22.18$ & $+22.18$ & $-12.74$ \\
\texttt{llama\_then\_gemma} & $+19.04$ & $+19.04$ & $-19.41$ \\
\texttt{llama\_then\_llama} & $+10.96$ & $+10.96$ & $-6.24$ \\
\texttt{llama\_then\_qwen} & $+18.07$ & $+18.07$ & $-7.55$ \\
\texttt{qwen\_then\_gemma} & $+24.85$ & $+24.85$ & $-21.11$ \\
\texttt{qwen\_then\_llama} & $+29.08$ & $+29.08$ & $-5.88$ \\
\texttt{qwen\_then\_qwen} & $+25.10$ & $+25.10$ & $-9.94$ \\
\bottomrule
\end{tabular}
\end{table}

\end{document}